\documentclass{article}

\usepackage{microtype}
\usepackage{graphicx}
\usepackage{subcaption}
\usepackage{booktabs} 

\usepackage{hyperref}

\usepackage[accepted]{icml2020}

\icmltitlerunning{Lifelong Learning using Eigentasks}

\usepackage{amsmath}
\usepackage{amssymb}
\usepackage{amsthm}
\usepackage{bbm}
\usepackage{multirow}
\usepackage{placeins}

\newcommand\Reals{\mathbb{R}}
\newcommand\Prob{\text{Prob}}
\newcommand\BT{\text{BT}}
\newcommand\FT{\text{FT}}
\newcommand\States{\mathcal{S}}
\newcommand\Actions{\mathcal{A}}

\begin{document}

\twocolumn[
\icmltitle{Lifelong Learning using Eigentasks:\\Task Separation, Skill Acquisition, and Selective Transfer}



%
\begin{icmlauthorlist}
\icmlauthor{Aswin Raghavan}{SRI}
\icmlauthor{Jesse Hostetler}{SRI}
\icmlauthor{Indranil Sur}{SRI}
\icmlauthor{Abrar Rahman}{SRI}
\icmlauthor{Ajay Divakaran}{SRI}
\end{icmlauthorlist}

\icmlaffiliation{SRI}{SRI International, Princeton, NJ, USA}

\icmlcorrespondingauthor{Aswin Raghavan}{aswin.raghavan@sri.com}

\icmlkeywords{Lifelong Learning, Reinforcement Learning, Continual Learning, Machine Learning, Starcraft}

\vskip 0.3in
]



\printAffiliationsAndNotice{}  

\begin{abstract}
We introduce the eigentask framework for lifelong learning. An eigentask is a pairing of a skill that solves a set of related tasks, paired with a generative model that can sample from the skill's input space. The framework extends generative replay approaches, which have mainly been used to avoid catastrophic forgetting, to also address other lifelong learning goals such as forward knowledge transfer. We propose a wake-sleep cycle of alternating task learning and knowledge consolidation for learning in our framework, and instantiate it for lifelong supervised learning and lifelong RL. We achieve improved performance over the state-of-the-art in supervised continual learning, and show evidence of forward knowledge transfer in a lifelong RL application in the game Starcraft~2.
\end{abstract}

\section{Introduction}
\label{sec:introduction}

The goal of lifelong learning \cite{chen_lifelong_2016-1, silver_lifelong_nodate} is 
to continuously learn a stream of machine learning tasks
over a long lifetime, accumulating knowledge and leveraging it to learn novel tasks
faster (positive forward transfer) \cite{fei_learning_2016} without forgetting the solutions to previous tasks (negative backward transfer). The learning problem is 
non-stationary and open-ended. The learner does not know the number or distribution
of tasks, it may not know the identity of tasks or when the task changes,
and it must be scalable to accommodate an ever-increasing
body of knowledge within a finite model. These characteristics make lifelong
learning particularly challenging for deep learning approaches, which
are vulnerable to catastrophic forgetting in the presence of non-stationary data. 


Many approaches to continual learning \cite{parisi_continual_2019,nguyen_variational_2018} and 
lifelong learning have been studied and applied to
computer vision \cite{hayes_memory_2018, liu_continual_2020} and reinforcement learning (RL) 
\cite{ammar_autonomous_nodate, tessler_deep_2016} among other areas. Recent work comprises many complementary 
approaches including learning a regularized parametric representation \cite{kirkpatrick_overcoming_2016,kirkpatrick_reply_2018}, 
transfer of learned representation \cite{lee_learning_2019}, 
meta-learning \cite{nagabandi_deep_2019}, neuromodulation \cite{masse_alleviating_2018}, 
dynamic neural network architectures \cite{li_learn_2019,rusu_progressive_2016}, and 
knowledge consolidation using memory \cite{van_de_ven_three_2019}. 

\begin{figure}[t]
\includegraphics[width=\columnwidth]{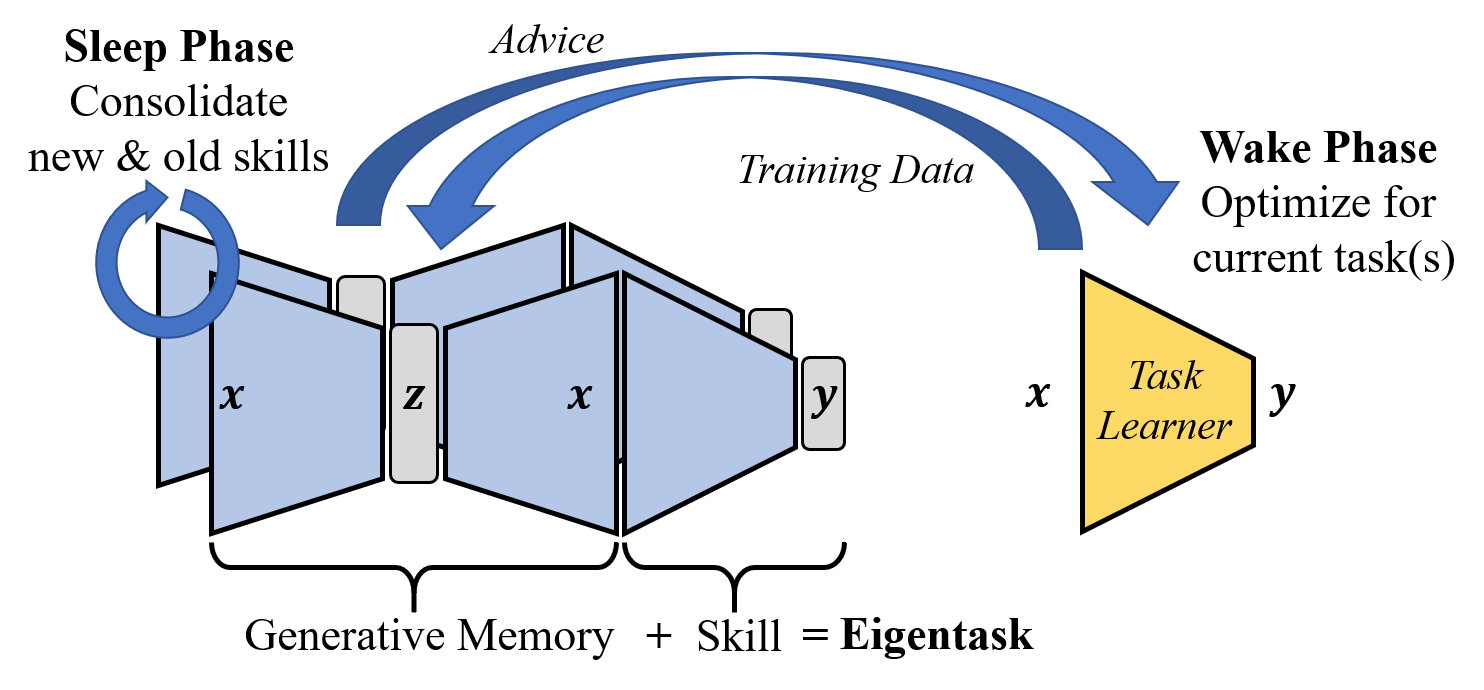}
\caption{Our approach is based on \emph{eigentasks}, which combine a skill with a generative model of its inputs. Our lifelong learning agents operate in a wake-sleep cycle, solving a stream of tasks during the wake phase, and consolidating new task knowledge during the sleep phase using generative replay.}
\vspace{-0.4cm}
\end{figure}

This paper advances the state of the art in the generative replay approach to lifelong learning
\cite{van_de_ven_three_2019,shin_continual_nodate}. 
In this approach, a generative model of the data distribution $p(x,y)=p(x)p(y|x)$ 
is learned and used for data augmentation i.e.,~replay data from old tasks when learning a new task. 
We refer to $p(x)$ as the generator (e.g., of images) and $p(y|x)$ as a skill (e.g., a classifier).
Prior work has focused on generative replay as a way to mitigate 
catastrophic forgetting, i.e.,~avoiding negative backward transfer.
However, forward transfer -- the ability to leverage knowledge to quickly adapt to a 
novel but related task -- is an equally important lifelong learning problem that
generative replay has not addressed. 
A structured form of replay is one of the functions of mammalian sleep  \cite{krishnan_biologically_2019,louie_temporally_2001}. 
Recent evidence in biology \cite{mcclelland_integration_2020} shows that 
only a few selected experiences are replayed during sleep,
in contrast to the typical replay mechanisms in machine learning. 
Our proposed approach bridges these gaps.


Our main contribution is a framework that combines generative memory with 
a set of skills that span the space of behaviors necessary to solve any given task. 
The framework consists of a set of generator-skill 
pairs that partition a stream of data into what we call \emph{eigentasks}. Each 
generator models a subset of the input space, and the corresponding skill encodes 
the appropriate outputs for the inputs in the generator's support set.
The likelihood of an input according to each generator is used to retrieve
the appropriate skill, making the eigentask model a content-addressable memory
for skills and enabling forward transfer.
Eigentasks can be seen as a combination of generative memory 
with mixture-of-experts (MoE) models \cite{makkuva_breaking_2018,tsuda_modeling_2020}. 
The MoE component facilitates forward transfer, while the generative component avoids
forgetting.  

We develop a concrete instantiation of eigentasks called Open World Variational Auto Encoder (OWVAE) 
that uses a set of VAEs \cite{kingma_auto-encoding_2014} as generators. OWVAE partitions data into eigentasks using 
out-of-distribution detection based on a likelihood ratio test in the latent space 
of the generator. We present a loss function for end-to-end 
learning of eigentasks that incorporates the losses of the generators and skills, 
weighted by the likelihood ratio. 
We show experimentally that OWVAE achieves superior performance 
comapred to state-of-the-art (SOTA) generative memory (GM) 
approaches on a new benchmark that contains a mix of MNIST and FashionMNIST datasets, 
and comparable performance in the splitMNIST benchmark. 
OWVAE's superior performance is attributed to task disentanglement \cite{achille_life-long_nodate} and confirmed 
visually by comparing the samples generated by each VAE. 



Our second contribution is a sampling strategy that improves the quality of
generative replay by using the confidence of the predictions output by the paired skills
to reject out-of-distribution samples. 
Our experiments show improved continual learning and reduction in forgetting when 
using rejection sampling as compared to accepting all generated examples. 

Our third contribution is a lifelong RL algorithm that leverages the  
OWVAE for exploration of new tasks. 
The OWVAE generates advice by recalling options \cite{sutton1999between} 
that are applicable to the current task, based on the eigentask partitioning of old tasks. 
This advice is incorporated into an 
off-policy learner whose behavior policy is a mixture of the options  
and the policy being learned.
We apply our lifelong RL algorithm to the challenging video game Starcraft~2 (SC2). 
On a sequence of ``mini-games'' \cite{vinyals2017starcraft} as tasks, our approach shows positive forward transfer when the 
new task is of the same type as one of the old tasks. In most mini-games we observe a jump start in the accumulated reward, and in one mini-game, forward transfer outperforms the single-task learner by 1.5x with 10x fewer samples due to better exploration.

\section{Background}
\label{sec:background}
In lifelong learning, an agent is faced with a never-ending sequence
of \emph{tasks}. Each task $i$ is defined by the tuple $T_i = (\mathcal{X}_i, \mathcal{Y}_i, P_i, L_i)$,
where $\mathcal{X}_i$ is the input space, $\mathcal{Y}_i$ is the output space,
$P_i \in \Prob(\mathcal{X}_i)$ is the input distribution, and
$L_i(\mathbf{x}, \mathbf{y}) : \mathcal{X}_i^* \times \mathcal{Y}_i^* \mapsto \Reals$ gives the
loss for outputting $\mathbf{y} = \{y_1, \ldots, y_k\}$ for inputs $\mathbf{x} = \{x_1, \ldots, x_k\}$. In this paper we consider the case where all the input and output
spaces are the same: $\mathcal{X}_i \equiv \mathcal{X}$, $\mathcal{Y}_i \equiv \mathcal{Y}$,
and our definition 
of a task is $T_i = (P_i, L_i)$.
We denote the (countably infinite) set of tasks by $\mathcal{T}$.
Since a task $T_i$ is totally characterized by its index
$i$, we use tasks and their indices interchangeably.

A task sequence is a sequence of tuples $((i_1, n_1), (i_2, n_2), \ldots)$
drawn from the task distribution $W$, where each $i_k \in \mathcal{T}$ is a task index and the corresponding
$n_k$ is the number of samples from $P_{i_k}$ that the agent sees before the transition
to the next task. 
The lifelong learning agent's objective is to learn a single hypothesis
$h : \mathcal{X} \mapsto \mathcal{Y}$ that minimizes the expected per-sample loss wrt the task distribution,
\begin{equation}
\mathcal{L}[h] = E_{(i, n) \sim W} E_{\mathbf{X}^n \sim P_i}[L_i(\mathbf{X}^n, h(\mathbf{X}^n))].
\label{eq:l2_obj}
\end{equation}

where $\mathbf{X}^n$ are $n$ instances sampled IID from $P$. 
Equation~\ref{eq:l2_obj} defines the optimal hypothesis wrt the task distribution.
Note that the task distribution $W$ is unknown to the agent, and $\mathbf{X}^n$ does not contain the task index.
Lifelong learning (Eq.~\ref{eq:l2_obj}) is strictly harder than multi-task learning, where
instances have known task IDs.


\subsection{Backward and Forward Transfer}
\label{sec:background:transfer}
The notion of knowledge transfer is central to evaluating success in lifelong learning.
In this paper, we approach lifelong learning in an episodic setting where the agent
learns in a series of learning epochs of fixed length (not necessarily aligned with task boundaries).
In this setting, we can define the key metrics
of forward and backward transfer in a manner that is agnostic to the task boundaries.
Let $N_i^j$ be the total
number of samples seen from task $i$ through the end of epoch $j$. We define a
loss restricted to tasks actually seen by the learner through epoch $j$ as
\begin{equation}
    \ell^j[h] = \sum_{i \in \mathcal{T}} \mathbbm{1}(N_i^j > 0) E_{X \sim P_i}[L_i(X, h(X))]
    \label{eq:l2_episodic_obj}
\end{equation}

Backward Transfer (BT) describes the difference in performance on old tasks
before and after one or more epochs of learning. 
Let $h^{j:k}$ be the hypothesis obtained after training \emph{sequentially} on the data
from epochs $j, j+1, \ldots, k$, where $h^j$ is shorthand for $h^{j:j}$.
The one-step backward transfer is,
\begin{equation}
    \BT(j) = \ell^{j-1}[h^{1:j}] - \ell^{j-1}[h^{1:j-1}].
    \label{eq:l2_bt}
\end{equation}
Negative BT corresponds to forgetting knowledge learned in previous episodes,
and positive BT could indicate successful knowledge consolidation between tasks.
Forward Transfer (FT) describes the difference in loss on new tasks
with and without training on previous tasks. 
\begin{equation}
    \FT(j) = (\ell^{j}[h^{1:j}] - \ell^{j-1}[h^{1:j}]) - (\ell^j[h^j] - \ell^{j-1}[h^j]) .
    \label{eq:l2_ft}
\end{equation}
The terms within parentheses restrict the loss to new tasks in the $j$th episode.
Positive FT indicates transfer of knowledge or skills between tasks, and typically
corresponds to jump start performance and lower sample complexity. 

\subsection{Reinforcement Learning}
\label{sec:background:rl}
The flexibility of our eigentask framework allows it to be applied in supervised learning (SL), unsupervised learning, and reinforcement learning (RL).
For simplicity, we describe lifelong RL in the finite Markov decision process (MDP) setting. 
An MDP is a tuple $(\States, \Actions, P, R)$, where $\States$ is a finite set of states, $\Actions$ is a finite set of actions, $P : \States \times \Actions \mapsto \Prob(\States)$ is the transition function, $R : \States \times \Actions \mapsto \Reals$ is the reward function. 
The objective is a policy $\pi : \States \mapsto \Prob(\Actions)$ that maximizes the value function $V^{\pi}(s) = E_{S_t \sim P, A_t \sim \pi}[\sum_{t = 1}^{\infty} \gamma^t R(S_t, A_t)]$. Tasks in lifelong RL are MDPs, with common state and action spaces, but $P$ and $R$ may differ. 
The loss is the negative return, $L(\mathbf{x}, \mathbf{y}) = -\sum_{(x, y) \in (\mathbf{x}, \mathbf{y})} R(x, y)$. 

\section{Eigentask Framework}
\label{sec:eigentasks}
Lifelong learning agents must balance plasticity vs. stability: improving at the
current task while maintaining performance on previously-learned tasks.
The problem of stability is especially acute in neural network models, where 
na\"{i}vely training an NN model on a sequence of tasks
leads to \emph{catastrophic forgetting} of previous tasks. A proven technique 
for avoiding forgetting is to mix data from all previous
tasks with data from the current task during training, thereby reverting
the streaming learning problem back to an offline learning problem.
The generative memory (GM) approach accomplishes this with a generative model of the data distribution $p(x,y)$ factorized
as a generator $p(x)$ over the input space $\mathcal{X}$ and a discriminator or skill $p(y|x)$
conditioned on the input $x$.
A skill can correspond to a classifier in supervised learning (SL)
or to a policy or option in RL.

To achieve selective transfer, we propose to use multiple generator-skill pairs to disentangle
streaming data into ``canonical tasks'' that we call \emph{eigentasks}.
Informally, eigentasks partition the joint input-output space such that all inputs within an
eigentask use the same skill. 
Eigentasks capture task similarity defined in terms
of the combination of generative and skill losses.
The use of multiple generators corresponds to a mixture model. Prior work on mixtures of VAEs
or GANs \cite{dilokthanakul_deep_2016,zhang_deep_nodate,rao2019continual} cluster inputs based on 
perceptual similarity alone or create one task per label. 
On the other hand, mixtures-of-experts
\cite{tsuda_modeling_2020} capture skill similarity alone.
The eigentask framework combines the MoE concept with generative replay,
to avoid forgetting and realize selective transfer. 

\subsection{Eigentask Definition}
\label{subsec:et}

Formally, an eigentask is a tuple $E = (g, f)$ comprising a generator $g(\epsilon)$
that defines a distribution over inputs $\mathcal{X}$ as a
function of random noise $\epsilon$,
and a skill $f : \mathcal{X} \mapsto \Prob(\mathcal{Y})$ that maps inputs to outputs.
An eigentask model $M = (\mathcal{E}_n, \tau)$
consists of a set of eigentasks $\mathcal{E}$ and a similarity function
$\tau: \mathcal{X} \mapsto\ \Prob(\mathcal{E})$ whose
output is a probability vector over the $n$ eigentasks. Typically, $g$, $f$ and $\tau$
are parameterized functions such as DNNs.
We write $\tau$ as a function of the
current input, but in general it may be a function of the entire input history.
We first describe the loss function for end-to-end learning of eigentasks in the offline setting.
The extension to the streaming section is discussed in the next section.
Given a dataset $\{X, Y\}$, the general loss function for eigentask learning is,
\begin{equation}
\small
\min \sum_{i=1}^n \tau_i(X) [\mathcal{L}_{gen}(g_i(\epsilon)|X) + \mathcal{L}_{disc}(f_i|g_i(\epsilon),X,Y)]
\label{eq:eigtask_loss}
\end{equation}
where $\mathcal{L}_{gen}$ and $\mathcal{L}_{disc}$ denote generative and discriminative losses for the
generator and skill, respectively. The loss in (\ref{eq:eigtask_loss}) is agnostic to the choice of
generator (e.g., VAE or GAN), skill (e.g. classifier, policy, options), and similarity function 
$\tau$. Note that the generator-skill pairs are independent, to avoid interference between
eigentasks. This is in contrast to recent approaches 
that learn a shared embedding across tasks \citep[e.g.][]{achille_life-long_nodate}.
In our experiments, $\mathcal{L}_{disc}$ is a cross-entropy loss
for classification problems and an RL loss such as the policy gradient for RL problems.

\begin{figure}[t]
    \centering
    \includegraphics[width=0.8\linewidth, keepaspectratio]{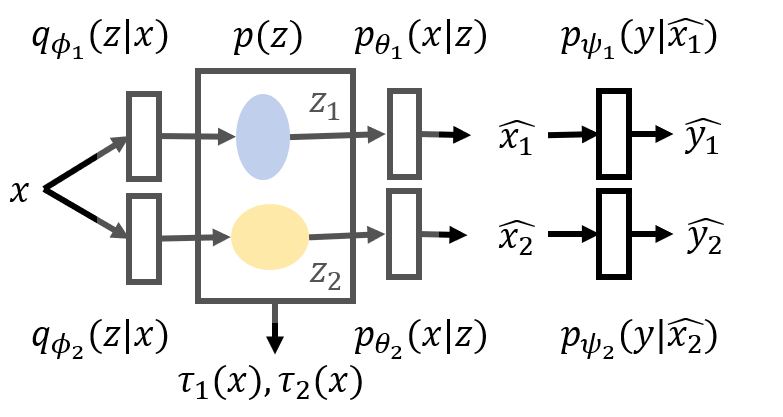}
\caption{An Open World Auto Encoder (OWVAE) with two eigentasks, showing the terms used in the loss function (Eq.~\ref{eq:owvae_loss}).}
    \label{fig:owvae}
\vspace{-0.5cm}
\end{figure}

We develop a concrete instantiation of eigentasks called Open World Variational Auto Encoder (OWVAE) 
that uses a set of VAEs \cite{kingma_auto-encoding_2014} as generators, with latent space denoted by
$z$, encoder $q_\phi(z|x)$, decoder $p_\theta(x|z)$ and prior $z \sim N(0, \mathbf{I})$. 
Figure~\ref{fig:owvae} shows an OWVAE with two eigentasks. 
We use reconstruction
error between the input and decoder output as the generative loss $\mathcal{L}_{gen}$. 
We propose to use a likelihood ratio test to define $\tau(x)$. \footnote{\cite{ren_likelihood_2019} developed a similar LR-test concurrently.}
We use the likelihood of decoder $i$ generating the observed data $x$ for some $z$,
that is approximated using encoding $z_i$.
\begin{equation}
    \tau_i(x) = \sigma\Big(\frac{p_{\theta_i}(x|z)}{\max_j p_{\theta_j}(x|z)}\Big) 
    \approx \sigma\Big(\frac{\Phi(z_i)}{\max_j \Phi_j(z_j)}\Big) \label{eq:tau_latent}
\end{equation}
where subscript $i$ denotes eigentask $i$, $z_i=q_{\phi_i}(z|x)$ is the encoding, $\Phi$ is the density of standard gaussian, $\sigma$ denotes softmax. The assumption is valid when the decoder weights are
the inverse of the encoder weights. 
The loss function for OWVAE is in Eq.~\ref{eq:owvae_loss}, 
\begin{equation} 
\min_{\theta, \phi, \psi} E_{x,y} [E_{\tau(x)} [\mathcal{L}_{VAE}(x; \theta, \phi) + \log p_\psi(y|\hat{x})]]	
\label{eq:owvae_loss}
\end{equation}
where $\mathcal{L}_{VAE}$ is the standard VAE loss,
\begin{equation*} \resizebox{.95\hsize}{!}{$\displaystyle
\mathcal{L}_{VAE}(x;\theta,\phi) = E_{q_\phi(z|x)} \log p_\theta(x|z) - D_{KL}(q_\phi(z|x)||p(z))$}.
\end{equation*}

%
%
%

At test time, given an input $x$ we calculate $\tau(x)$ by computing a forward pass
through the encoder. We sample the index of the decoder and skill to use
according to the categorical distribution given by $\tau(x)$.
In the simplest case, we pass the decoder output as
the input to the skill. In our experiments, we observed that using the mid-level
features (the encoder features before projecting to latent space) led to the best
accuracy of the skill, matching an observation by \citet{van_de_ven_generative_2018}.
Note that the above sampling method is conditional on $x$, and that the OWVAE does
not allow direct sampling from the learned mixture because the mixing coefficient $\tau(x)$ depends on the input.
Generative replay can be used to learn an OWVAE incrementally but requires direct sampling of old tasks. 
Section \ref{sec:lifelong_sl} proposes a strategy for direct sampling.

\begin{algorithm}[t]
   \caption{General Wake-Sleep Cycle}
   \label{alg:wakesleep}
\begin{algorithmic}
   \STATE {\bfseries Input:} OWVAE $M$, buffer $B$, \# sleep iterations $N$
   \STATE Initialize $M$; set $B$ to empty.\\
	   \WHILE{True}
	   \STATE Initialize task learner $T$
	   \REPEAT[Wake Phase]
		   \STATE{Classification (Section \ref{sec:lifelong_sl}): store new instance in buffer}
		   \STATE{RL (Section \ref{sec:lifelong_rl}): Update $T,B$ with Alg \ref{alg:owvae_explore}}
	   \UNTIL{$B$ is full}\\
	   \STATE{Create copy $M_2=M$}
	   \FOR[Sleep Phase]{$N$ iterations} 
		   \STATE{Fetch batch $b$ from $B$}
			 \STATE{Generate replay $b_M \sim M_2$ using Alg \ref{alg:owvae_sampling}}
			 \STATE{Update $M$ using Eq. \ref{eq:owvae_loss} on $b \cup b_M$}
		 \ENDFOR\\
		 \STATE Set $B$ to empty
	 \ENDWHILE
\end{algorithmic}
\end{algorithm}

\subsection{The Wake-Sleep Cycle}
Our lifelong learning agents operate in a wake-sleep cycle, solving a stream of tasks during the wake phase, and consolidating new task knowledge during the sleep phase using generative replay.
In the wake phase, the learner's goal is to maximize FT wrt correctly outputs on incoming inputs. 
In supervised learning that might mean simply outputting the correct label according to the new task,
while in reinforcement learning the agent needs to explore the new task and maximize reward. 
During the wake phase, the learner converts the streaming input to batched data, 
by storing new task examples in a short-term buffer along with any
intermediate solutions of wake phase learning. Periodically, the learner enters a sleep phase, where the
objective is memory consolidation over all tasks with minimal negative BT. 
We use generative replay to incorporate new task batches in to an eigentask model that is continuously updated.
A general wake-sleep cycle is shown in Algorithm \ref{alg:wakesleep} where sleep phase is activated whenever the buffer is full. 
We describe instantiations of the wake-sleep cycle for supervised learning and RL in the next two sections.

\section{Lifelong Supervised Learning}
\label{sec:lifelong_sl}



\textbf{Wake phase}: In this work, we use a trivial wake phase for supervised classification. 
We simply store the new task examples (instance and label) to the buffer. 
The OWVAE could be leveraged in the wake phase, e.g. augment 
new task data with selective replay of similar tasks, 
or using the OWVAE skills as hints for knowledge distillation.
For example, hints led to positive FT when the new task was a 
noisy version of old tasks (noise added to labels and pixels).
We have not yet investigated these possibilities completely as they are specific to the scenario. 
Algorithms for learning from streaming data 
\citep[e.g.][]{hayes_memory_2018, smith2019unsupervised} can be used to 
update the task learner.

\textbf{Sleep phase}:  
As mentioned in Section \ref{subsec:et}, sampling from an OWVAE is 
conditional on input $x$ because the mixing coefficients $\tau(x)$ are a function of $x$. 
Some eigentasks may have received little or no training (e.g. when old tasks are few and similar). 
These untrained generators will generate noise that must not be used to augment new task data. 
To mitigate this problem, we developed a rejection sampling strategy 
using the confidence of the skill associated with the generator.
Algorithm \ref{alg:owvae_sampling} shows the sampling strategy for OWVAE. 
We reject a sample $(x, y)$ if the confidence of the associated skill
is below a threshold $\delta$. 
We reject over-represented labels and generate label-balanced replay. 
These two refinements to the sampling process were 
critical in achieving high accuracy on continual learning benchmarks.
In addition, rejection sampling improved the accuracy of the basic GM approach as well.

\begin{algorithm}[t]
   \caption{Rejection Sampling from OWVAE}
   \label{alg:owvae_sampling}
\begin{algorithmic}
   \STATE {\bfseries Input:} OWVAE $M$, batch size $K$, threshold $\delta$
   \STATE Initialize deque $D[y]$ for each label $y$ of size $K$
   \REPEAT
	   \FOR{Eigentask $i$ in $M$}
   	  \STATE Generate $z_i \sim N(0,\mathbf{I})$, $x_i \sim p_{\theta_i}(x|z_i)$
	   	\STATE $y_i=\arg\max p_{\psi_i}(y|x_i)$, $\text{conf}_i=\max p_{\psi_i}(y|x_i)$
		  \STATE \textbf{if} $\text{conf}_i > \delta$: Push $x_i, y_i$ to $D[y_i]$
	   \ENDFOR
   \UNTIL{Each $D[y]$ is full or MAX\_TRIES}
   \STATE Return $\bigcup_y D[y]$
\end{algorithmic}
\end{algorithm}

\section{Lifelong Reinforcement Learning}
\label{sec:lifelong_rl}
In the lifelong RL setting, tasks are MDPs (Section~\ref{sec:background:rl}), the eigentask skills are policies, and the associated generators generate states where the policies should be applied. 

\textbf{Wake Phase}: 
One of the key determinants of RL performance is the efficiency of exploration. 
Without any prior knowledge, RL algorithms typically explore randomly in the early stages of learning. 
Our approach (see Algorithm \ref{alg:owvae_explore}) is to use the skill 
corresponding to the most relevant eigentask to aid in exploration.
An off-policy RL algorithm is used for training because the exploration policy is defined by these skills. 
The behavior policy is a mixture of the target policy and the mixture of skills induced by the OWVAE's $\tau$-function.
Let $\pi$ denote the target policy and let $p_{\psi_i}$ be the $i$th skill. 
Given a state $s$, the behavior policy $b(s)$ is, 
\begin{equation}
b(s) = \left\{\begin{array}{ll}
  \pi(s) & \text{w.p. } 1 - \mu(t) \\
  p_{\psi_i}(s) & \text{w.p. } \mu(t) \tau(s)_i \text{ each eigentask $i$}
\end{array}\right. \label{eq:rl_mixing}
\end{equation}
where $\mu(t) \in [0,1]$ is a ``mixing'' function that decays over time so that 
eigentask usage is gradually reduced and replaced by $\pi$. 
We use importance weighting as implemented by the off-policy actor-critic algorithm VTrace \citep{espeholt2018impala}, 
but our approach is compatible with any off-policy algorithm. 
In the last step of the wake phase, trajectories from the target policy are stored in the buffer, 
to be later consolidated in the sleep phase. 

\begin{algorithm}[t]
   \caption{Exploration using OWVAE for lifelong RL}
   \label{alg:owvae_explore}
\begin{algorithmic}
   \STATE {\bfseries Input:} 
   \begin{small} OWVAE $M$, Buffer $B$, Off-policy learner $A$, MDP $\mathbb{M}$ \end{small}
   \STATE Initialize policy $\pi$, $\mu(t)$
   \REPEAT
   	\STATE $s=$ state from $\mathbb{M}$, get $\tau(s)$ from OWVAE using Eq. \ref{eq:tau_latent}
   	\STATE Sample action $a$ using $b(s)$ and $\tau(s)$ as in Eq. \ref{eq:rl_mixing}
   	\STATE Execute $a$ in $\mathbb{M}$. Observe next state $s'$ and reward $r$
   	\STATE Add $(s,a,s',r)$ to RL training set
   	\STATE Update $\pi$ using $A$. Decrease $\mu(t)$
	\UNTIL{Sample budget reached}
	\STATE Reset $\mathbb{M}$ to initial state
	\STATE Execute $\pi$ to generate set of $(s,a,s',r)$
	\STATE Add $(s,a,s',r)$ to buffer $B$
\end{algorithmic}
\end{algorithm}

\textbf{Sleep Phase}:
In the sleep phase, the goal is to consolidate the final target policy $\pi$ into the eigentask skills $p_{\psi_i}$. 
Our approach is based on policy distillation \citep{rusu2015policy}, that transforms the problem to a supervised learning problem.
The sleep phase proceeds in the same manner as for supervised learning tasks (Section~\ref{sec:lifelong_sl}). 

\section{Experiments}
\label{sec:experiments}
We validate the idea of eigentasks on unsupervised, supervised classification and RL tasks.
We show experimentally that 
OWVAE achieves superior performance 
comapred to state-of-the-art (SOTA) generative memory (GM) 
approaches RtF \citep{van_de_ven_three_2019} and DGR \cite{shin_continual_nodate} 
on a new benchmark that contains a mix of MNIST and FashionMNIST datasets, 
and comparable performance in the splitMNIST benchmark for continual learning. 
OWVAE's superior performance is attributed to 
task disentanglement \cite{achille_life-long_nodate} and confirmed 
visually by comparing the samples generated by each VAE. 
We demonstrate our lifelong RL algorithm on the
Starcraft 2 (SC2) mini-games benchmark \cite{vinyals2017starcraft}.
We demonstrate that our approach compares favorably to the baselines of single-task and multi-task
RL.

\subsection{Illustration: synthetic problem}
\label{sec:synthetic}
In this section, we illustrate the OWVAE model on a synthetic but challenging problem
for current GM approaches. The problem is inspired
by the Wisconsin Card Sorting task \cite{tsuda_modeling_2020}, where the 
tasks cannot be distinguished by ``perceptual similarity'', but can be distinguished
by ``skill similarity'', so a mixture-of-experts model can solve the tasks \cite{tsuda_modeling_2020}. 
We test whether an OWVAE can separate the tasks and achieve a high accuracy.

\begin{figure*}
\centering
\hfill
\begin{subfigure}[t]{0.28\textwidth}
\includegraphics[width=\textwidth, keepaspectratio]{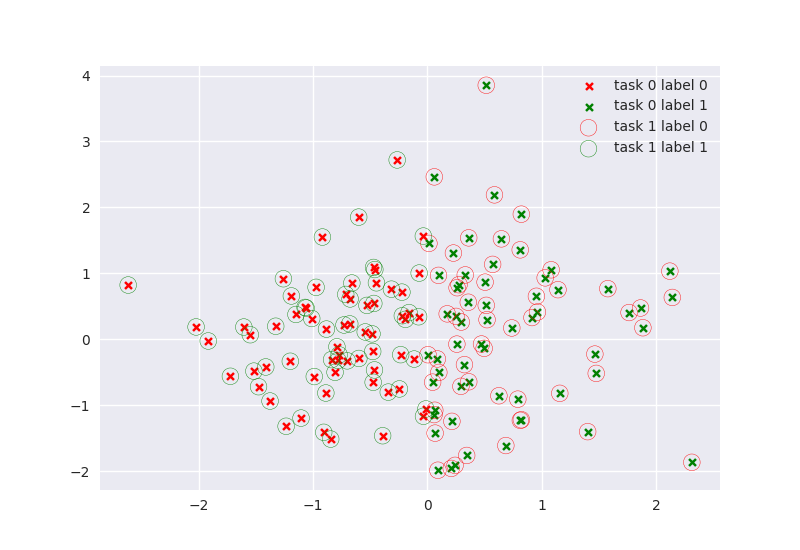}
\caption{Setup of the synthetic problem with conflicting tasks.}
\label{fig:wisc2d_data}
\end{subfigure}
\hfill
\begin{subfigure}[t]{0.28\textwidth}
\includegraphics[width=\textwidth,keepaspectratio]{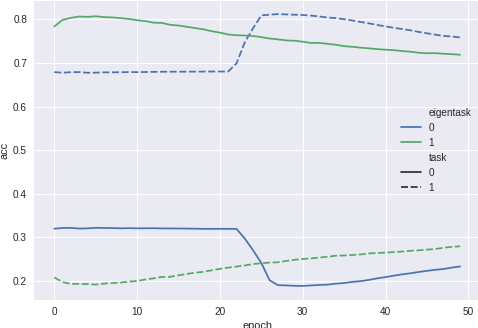}
\caption{Accuracy with OWVAE(2): both tasks can be separated and learned.}
\label{fig:wisc2d_accuracy}
\end{subfigure}
\hfill
\begin{subfigure}[t]{0.28\textwidth}
\includegraphics[width=\textwidth, keepaspectratio]{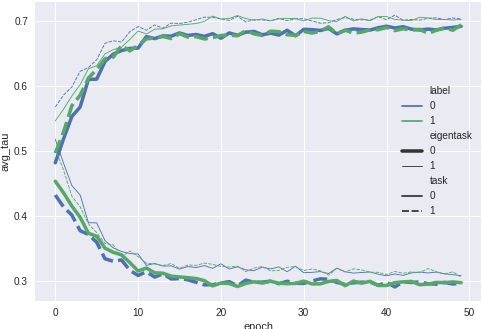}
\caption{Tracking $\tau$ shows correctly identified eigentasks.}
\label{fig:wisc2d_tau}
\end{subfigure}
\hfill
\caption{Illustration on synthetic problem (Section~\ref{sec:synthetic}): OWVAE(2) learns two conflicting tasks with no perceptual dissimilarity.}
\label{fig:wisc2d}
\end{figure*}

Consider two binary classification tasks whose input space is an isotropic Gaussian
in two dimenstions, and the labels for the two tasks are flipped, e.g. task-0 labels
are given by $(X \geq 0)$, and task-1 labels are $(X < 0)$. The data distribution
is shown in Figure \ref{fig:wisc2d_data}. Any classifier that achieves an accuracy of $p \in [0,1]$
on task-0 must have accuracy $1-p$ on task-1, and thus average accuracy of $0.5$.
Figure \ref{fig:wisc2d_accuracy} shows that an OWVAE with two eigentasks is able to 
achieve high accuracy ($>0.7$). Figure \ref{fig:wisc2d_tau} shows the
reason: a meaningful $\tau$ has been learned that separates
the tasks successfully into two eigentasks. Interestingly, the learned eigentasks
are $(X \geq 0)$ and $(X < 0)$, i.e. label-0 of task-0 is combined with label-1 of task-1
within one eigentask, and label-1 of task-0 is combined with label-0 of task-1 within another eigentask. 
In contrast, current GM methods will not work 
because learning either task causes forgetting of the other.

\subsection{Continual Learning for Supervised Classification}
\label{sec:mnist_experiments}
We use the class-incremental learning (Class-IL) setting introduced in \cite{van_de_ven_three_2019}.
In this setting, new classes or groups of classes are presented incrementally to the
learning alorithm.
We use the standard splitMNIST problem and compare to the SOTA.
In splitMNIST, the MNIST dataset is split into five tasks with each task having two of the original classes.
Further, we create a new benchmark combining MNIST and Fashion-MNIST datasets.
The combined MNIST and Fashion-MNIST classes are split evenly into ten tasks. 
Each task introduces two new classes, one MNIST digit and one fashion article.
In this new benchmark, we establish a new SOTA by showing that current replay based approaches are inferior to OWVAE.

In both benchmarks, each new task is trained for 500 iterations with a batch size of 32. 
All compared SOTA approaches use a two layer perceptron with 650 neurons for the encoder and decoder, and a latent
dimension of $100$. We compare OWVAE against eight continual learning approaches spanning regularization, replay, and replay with
exemplars (Table \ref{tab:my-table}). In addition, we show 
two baselines of single-task learner (lower bound - only learns on current task)
and offline multi-task learner (upper bound - knows all tasks).
Training is done on the standard train sets and results are reported
on the standard test sets.

The OWVAE uses two eigentasks, each with the same architecture as SOTA but with 400 neurons only, 
so that the OWVAE has the same number of parameters as SOTA. Within the OWVAE, 
the inputs to the skill are the mid-level features of the
corresponding encoder, i.e.~activations in the last layer of the encoder. 
As in Section \ref{sec:lifelong_sl}, no wake phase is used and sleep phase 
uses 500 iterations with a batch size of 32, and the threshold $\delta$ is set to $0.5$ (as in Alg. \ref{alg:owvae_sampling}).
To study the impact of rejection sampling,
we perform an ablation study over different augmentation strategies: (1) BaseAug: all examples
generated during replay are accepted, (2) BAug: rejection sampling to create label-balanced replay, 
(3) VAug: rejecting low confidence examples, and 
(4) VBAug: combining (2) and (3) (as in Alg \ref{alg:owvae_sampling}). 

\begin{table}[t]
    \centering
    \small
    \begin{tabular}{llcc}
    \hline
    Approach & Method & D1 & D2 \\ \hline \hline
    \multirow{2}{*}{Baselines}      & \multicolumn{1}{l}{None - lower bound} & \multicolumn{1}{c}{19.90} & \multicolumn{1}{c}{10.22} \\
                                    & \multicolumn{1}{l}{Offline - upper bound} & \multicolumn{1}{c}{97.94} & \multicolumn{1}{c}{90.89} \\\hline
    \multirow{3}{*}{Regularization} & \multicolumn{1}{l}{EWC} & \multicolumn{1}{c}{20.01} & \multicolumn{1}{c}{10.00} \\
                                    & \multicolumn{1}{l}{Online EWC} & \multicolumn{1}{c}{19.96} & \multicolumn{1}{c}{10.00} \\
                                    & \multicolumn{1}{l}{SI} & \multicolumn{1}{c}{19.99} & \multicolumn{1}{c}{10.00} \\\hline
    \multirow{4}{*}{Replay}         & \multicolumn{1}{l}{LwF} & \multicolumn{1}{c}{23.85} & \multicolumn{1}{c}{10.07} \\
                                    & \multicolumn{1}{l}{DGR} & \multicolumn{1}{c}{90.79} & \multicolumn{1}{c}{73.36} \\
                                    & \multicolumn{1}{l}{DGR x2} & \multicolumn{1}{c}{91.83} & \multicolumn{1}{c}{65.82} \\                                    
                                    & \multicolumn{1}{l}{DGR+distill} & \multicolumn{1}{c}{91.79} & \multicolumn{1}{c}{72.40} \\
                                    & \multicolumn{1}{l}{DGR+distill x2} & \multicolumn{1}{c}{94.01} & \multicolumn{1}{c}{67.37} \\\hline
                                    & \multicolumn{1}{l}{RtF} & \multicolumn{1}{c}{92.56} & \multicolumn{1}{c}{61.15} \\
                                    & \multicolumn{1}{l}{RtF x2} & \multicolumn{1}{c}{92.86} & \multicolumn{1}{c}{61.41} \\\hline
    \multirow{1}{*}{Replay+Exemplars}  & \multicolumn{1}{l}{iCaRL} & \multicolumn{1}{c}{94.57} & \multicolumn{1}{c}{82.69} \\\hline
    \multirow{6}{*}{Replay+Eigentask}    & \multicolumn{1}{l}{ET1-BaseAug} & \multicolumn{1}{c}{87.68} & \multicolumn{1}{c}{69.29} \\
                                    & \multicolumn{1}{l}{ET1-BAug} & \multicolumn{1}{c}{90.99} & \multicolumn{1}{c}{74.11} \\
                                    & \multicolumn{1}{l}{ET1-VAug} & \multicolumn{1}{c}{87.33} & \multicolumn{1}{c}{63.34} \\
                                    & \multicolumn{1}{l}{ET1-VBAug} & \multicolumn{1}{c}{90.69} & \multicolumn{1}{c}{77.43} \\\cline{2-4}
                                    & \multicolumn{1}{l}{ET2-BaseAug} & \multicolumn{1}{c}{88.93} & \multicolumn{1}{c}{57.91} \\
                                    & \multicolumn{1}{l}{ET2-BAug} & \multicolumn{1}{c}{\textbf{91.27}} & \multicolumn{1}{c}{69.95} \\
                                    & \multicolumn{1}{l}{ET2-VAug} & \multicolumn{1}{c}{82.08} & \multicolumn{1}{c}{69.55} \\
                                    & \multicolumn{1}{l}{ET2-VBAug} & \multicolumn{1}{c}{90.25} & \multicolumn{1}{c}{\textbf{76.81}} \\\hline
    \end{tabular}
    \caption{Average test accuracy over all tasks on splitMNIST (D1) and split(MNIST+FashionMNIST) (D2) benchmarks. ET1 and ET2 denote
    the number of eigentasks in an OWVAE model. Methods compared: EWC \cite{kirkpatrick_overcoming_2016}, Online EWC \cite{schwarz2018progress}, SI \cite{zenke2017continual}, LwF \cite{li_learning_2016}, DGR\cite{shin_continual_nodate}, RtF \cite{van_de_ven_three_2019},and iCaRL \cite{Rebuffi_2017_CVPR}. The variants denoted x2 have the same number of parameters as ET2.}
    \label{tab:my-table}
    \vspace{-0.45cm}
\end{table}
The average accuracies over all tasks at the end of training are shown in Table \ref{tab:my-table}.
As observed in prior work, the class-IL setting is hard for the regularization 
approaches like EWC, as well as LwF; their performance
is very low, comparable to the single task learner (they learn and immediately forget each task).
Our approach (Replay+Eigentasks) has accuracy comparable to other
replay-based methods (except LwF) on splitMNIST. However, on split(MNIST+FashionMNIST),
our approach has a higher accuracy (about 4\% higher). Unsurprisingly, using
exemplars within replay improves accuracy on both benchmarks. 
Exemplars could be integrated into OWVAE	 in future work.

The ablation study shows that VBAug (as in the combination of rejection using both confidence and label) 
leads to the most improvement (2-3\% on SplitMNIST, 19\% on split(MNIST+FashionMNIST)). 
The combination VBAug performs better than BAug, BaseAug and VAug, 
whereas VAug by itself seems to decrease the performance wrt BaseAug. 
Interestingly, VBAug and BAug improved the performance of the 
basic GM approach (as in BaseAug vs ET1: 2-3\% improvement on splitMNIST, 5-7\% improvement on split(MNIST+FashionMNIST)).
A detailed breakdown of accuracy per-task over time is shown in the appendix. 

Figure~\ref{fig:reconstruction_fm} shows the task separation learned by OWVAE. 
The figure visualizes VAE reconstructions for each task.
It shows that the first eigentask has learned all the MNIST digits and able to reconstruct them, 
whereas the second eigentask has learned all the fashion articles. 
The first eigentask has also learned some fashion articles whereas
the second eigentask has not learned any digits. The blurry and noisy images 
are removed from replay by our rejection sampling strategy. 

\begin{figure}[t]
    \centering
    \includegraphics[width=0.9\linewidth, keepaspectratio]{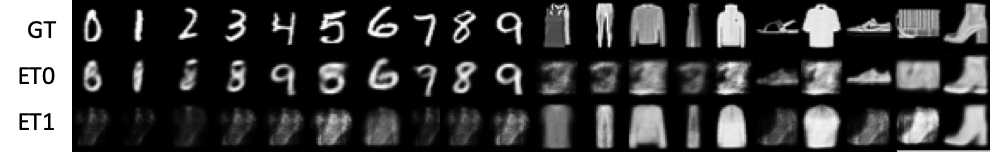}
    \caption{Split(MNIST+FashionMNIST): Image reconstruction by OWVAE(2) showing task separation. Top row: Ground truth. Middle and Bottom: reconstruction by first and second eigentask.}
	\label{fig:reconstruction_fm}
	\vspace{-0.4cm}
\end{figure}

\subsection{Starcraft 2}
\label{sec:sc2}
We use the Starcraft~2 learning environment (SC2) \citep{vinyals2017starcraft}.
SC2 is a rich platform in which diverse RL tasks can be implemented.
We used their ``mini-games'' 
as the task set for our experiments.
Our policy architecture is a slight modification of their FullyConv architecture. 
We use the 17 64x64 feature maps extracted by SC2LE. 
\subsubsection{Eigentask Learning}
In the OWVAE, we use only two feature maps namely 
``unit type'' (identity of the game unit present in each pixel) 
and ``unit density'' (average number of units per pixel).
An example of these feature maps is shown in the appendix. 

\begin{figure}
\centering
\includegraphics[width=0.6\linewidth,keepaspectratio]{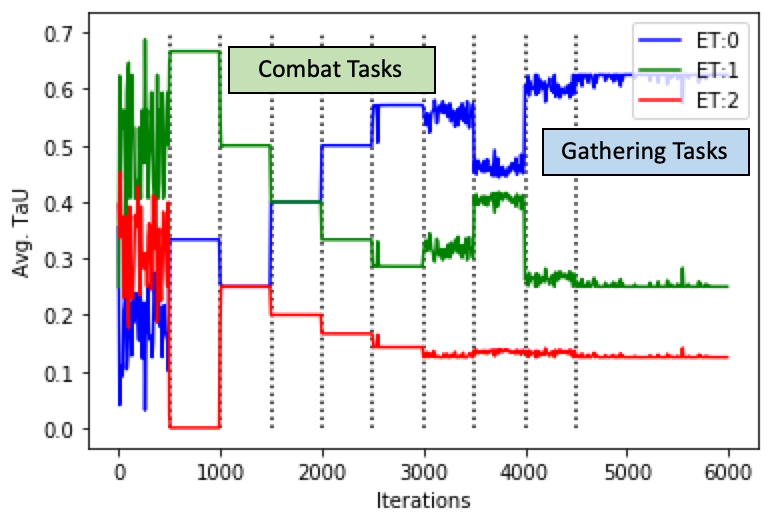}
\caption{Continual Unsupervised Learning of a OWVAE(3) in SC2 mini-games: Variation of $\tau$ over iterations and learned task grouping. Each mini-game is observed for 500 iterations.}
\label{fig:sc2_eigentasks}
\vspace{-0.4cm}
\end{figure}

The eigentasks for SC2 are learned in an unsupervised and offline manner.
We first collected a dataset by executing a random policy in each mini-game and recording the feature maps per frame.
Whenever possible, we also collected similar data by running a scripted agent.
We then trained an OWVAE incrementally with three eigentasks in a fully unsupervised manner. 
The setup is similar to the continual learning experiment for splitMNIST 
(each mini-game is seen for 500 iterations etc.). 
SC2 mini-games are perceptually different, so the OWVAE is able to separate the tasks into meaningful eigentasks 
despite the unsupervised training. 
As shown in Figure \ref{fig:sc2_eigentasks}, by looking at the relative values of the OWVAE $\tau$-function, 
we see that the first eigentask grouped all combat tasks together, 
but incorrectly grouped a navigation task (possibly due to the unsupervised training). 
The second eigentask learned the BuildMarines task alone, whereas the 
third eigentask grouped the resource gathering tasks together.
For each task type, we observe one eigentask clearly dominating the $\tau$ values, while no single eigentask dominates always.
These task groupings can be confirmed by looking at the VAE reconstructions shown in appendix.

\begin{figure*}
\centering
\begin{subfigure}[t]{0.32\textwidth}
\includegraphics[width=\textwidth]{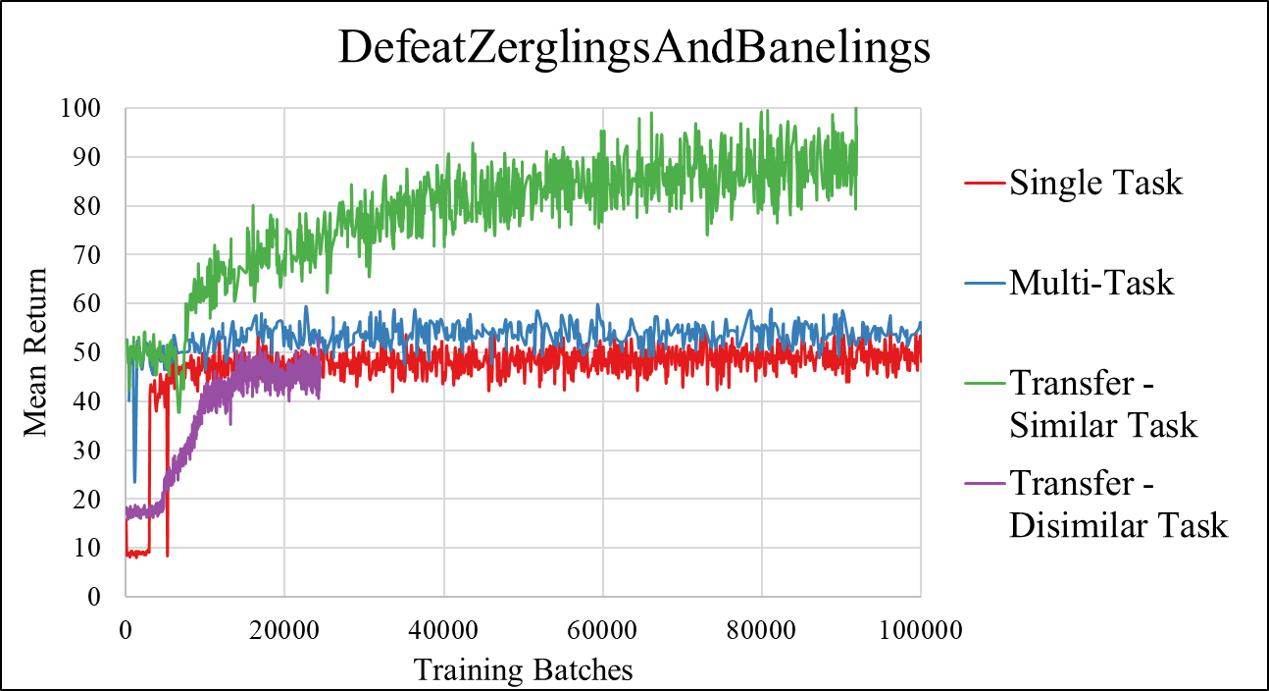}
\caption{}
\label{fig:sc2-forward-transfer:defeatzerglingsandbanelings}
\end{subfigure}
\hfill
\begin{subfigure}[t]{0.32\textwidth}
\includegraphics[width=\textwidth]{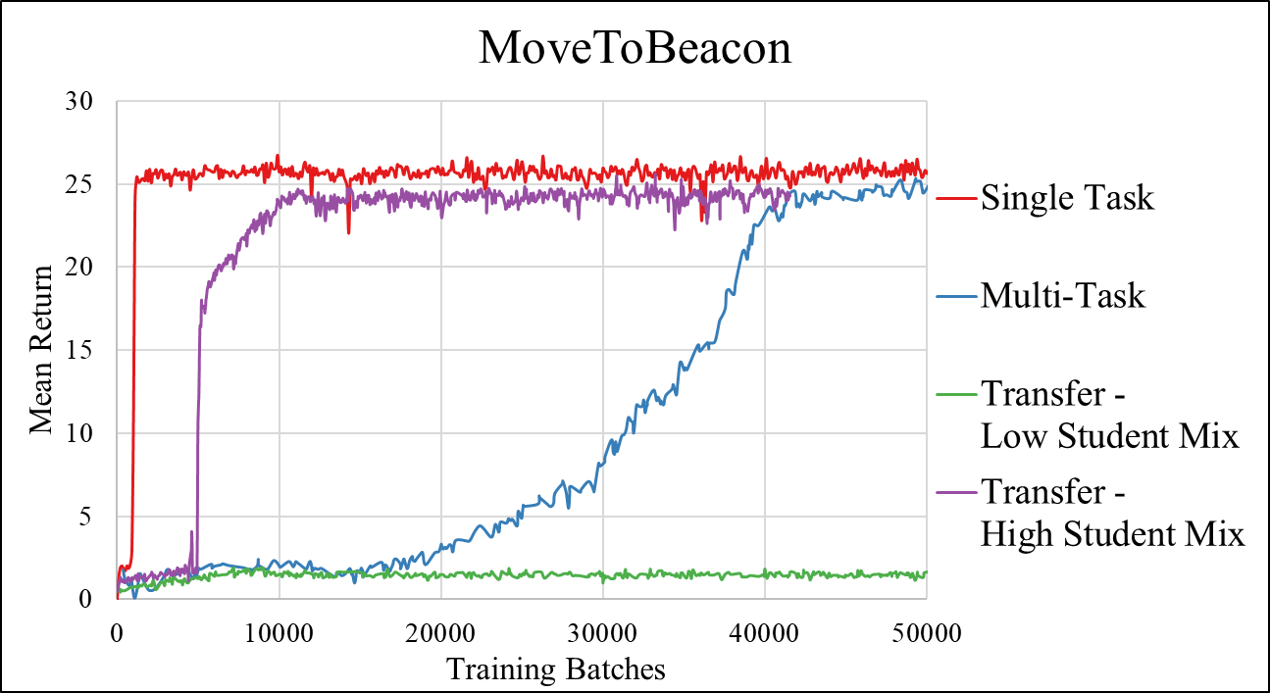}
\caption{}
\label{fig:sc2-forward-transfer:movetobeacon}
\end{subfigure}
\hfill
\begin{subfigure}[t]{0.32\textwidth}
\includegraphics[width=\textwidth]{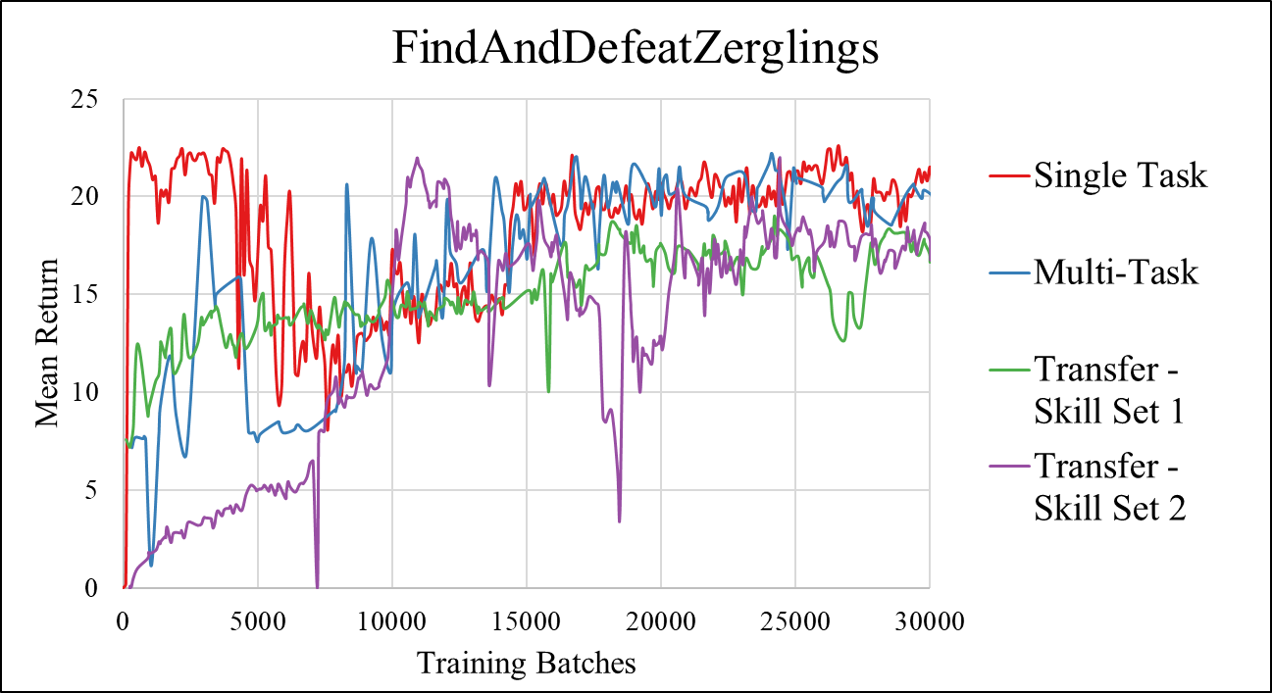}
\caption{}
\label{fig:sc2-forward-transfer:findanddefeatzerglings}
\end{subfigure}
\caption{Forward transfer in Starcraft~2 mini-games.}
\label{fig:sc2-forward-transfer}
\end{figure*}

\subsubsection{Forward Policy Transfer}
We focus on the wake phase of our lifelong RL algorithm (Alg \ref{alg:owvae_explore}),  
and examine forward transfer due to efficient exploration. 
We manually set the OWVAE skills selected from the set of trained single-task policies. 
These policies are separated into groups, based on the task separation 
observed from the unsupervised OWVAE training above. Each skill is assigned a policy from the corresponding group. 
Training uses the VTrace learning rule \citep{espeholt2018impala} 
in an A2C implementation heavily adapted from code published by \citet{ring2018reaver}. 
We examined transfer from skills from source tasks that are either similar or dissimilar to the target task (Table~\ref{tab:sc2-task-groups}). The main results with forward transfer are summarized in Figure~\ref{fig:sc2-forward-transfer}. The plots include two baselines: the performance of a single task policy trained on the target task, and a multi-task policy trained on batches containing experience from all 6 tasks. In order to demonstrate efficient exploration, the vertical axis shows mean per-episode return obtained by the behavior policy during training.

\begin{table}  \begin{center}
\begin{tabular}{l|l}
\textbf{Category} & \textbf{Tasks} \\
\hline
Combat & DefeatRoaches, \\
       & DefeatZerglingsAndBanelings \\
\hline
Navigation & MoveToBeacon, CollectMineralShards \\
\hline
Hybrid & FindAndDefeatRoaches \\
\hline
Economy & CollectMineralsAndGas
\end{tabular}
\caption{Categories of SC2 tasks. Tasks in the same category are considered ``similar'' in our experiments.}
\vspace{-0.5cm}
\label{tab:sc2-task-groups}
\end{center} \end{table}

Experiment (\ref{fig:sc2-forward-transfer:defeatzerglingsandbanelings}) examines transfer to the DefeatZerglingsAndBanelings task. In the Transfer-Similar Task condition, the OWVAE skills are policies trained on CollectMineralShards and DefeatRoaches. 
The DefeatRoaches task is another combat task, and thus is similar to the target task. 
In the Transfer-Disimilar Task condition, the OWVAE skills are CollectMineralShards and MoveToBeacon. 
In the Transfer-Similar condition, our approach yielded both good performance from the start of training and substantially better asymptotic performance. 
Interestingly, our approach even surpassed the asymptotic performance of single-task and multi-task learning, 
clearly showing the impact of better exploration transferred through the OWVAE skills. 
Furthermore, the asymptotic performance is also better than the best published 
performance for the FullyConv policy architecture \citep{vinyals2017starcraft} by about 
1.5x while using 10x fewer RL iterations. 
In the Transfer-Dissimilar condition, our approach still resulted in better initial 
performance than single task training, but converged to the asymptotic performance of 
the single task policy at a slower rate. 

In experiment (\ref{fig:sc2-forward-transfer:movetobeacon}), we study transfer to the MoveToBeacon task. 
Because of the way we trained the OWVAE, the MoveToBeacon task gets clustered with two Combat tasks, and not with the more similar CollectMineralShards task. 
As a result, the OWVAE $\tau$ function selects a combat skill to use for transfer to MoveToBeacon. 
When we use the default schedule for the behavior policy mixing rate $\mu(t)$, 
transfer from the inappropriate skill hinders learning. 
However, using a different mixing schedule that gives more weight to the target policy 
allows the agent to overcome the effect. 

Finally, experiment (\ref{fig:sc2-forward-transfer:findanddefeatzerglings}) investigates transfer to the FindAndDefeatZerglings task. This is an interesting target task because it combines elements of Combat and Navigation tasks, but is not highly similar to either of those categories. We evaluated two different skill sets (either CollectMineralShards+DefeatRoaches or CollectMineralShards+MoveToBeacon) but transfer had no clear positive or negative effect for this target task.

\section{Discussion and Future Work}
\label{sec:discussion}
We introduced the eigentask framework for lifelong learning, which combines generative replay with mixture-of-experts style skill learning. We use the framework in a wake-sleep cycle where new tasks are solved in the wake phase and experiences are consolidated into memory in the sleep phase. We applied it to both lifelong supervised learning and RL problems. We developed refinements to the standard generative replay approach to enable selective knowledge transfer. Combined with the rejection sampling trick, we  
achieved SOTA performance on continual learning benchmarks. In lifelong RL, we demonstrated successful forward transfer to new tasks in Starcraft~2, and exceeded the best published performance on one of the Starcraft~2 tasks. 

Our immediate goal in future work is to close the wake-sleep loop in lifelong RL. We have demonstrated success for components of the approach, but not for the full eigentask framework. We are interested in adding change-point detection to improve on the likelihood ratio test, and hierarchical eigentasks that could be more compact and more efficient. Finally, we want to incorporate task similarity measures that account for history, to separate RL tasks that have similar observations but different dynamics.

\FloatBarrier
\section*{Acknowledgements}
This material is based on work supported by the Lifelong Learning Machines (L2M) program of
the Defense Advanced Research Projects Agency (DARPA) under contract HR0011-18-C-0051.
Any opinions, findings and conclusions or recommendations expressed in this material are those of the authors and do not
necessarily reflect the views of DARPA, the Department of Defense or the U.S. Government. 
\nocite{yosinski_how_2014, huszar_note_2018, wilson_reactivation_1994,diekelmann_memory_2010, he_exemplar-supported_nodate, ramapuram_lifelong_2017}
\FloatBarrier
\bibliography{ref}
\bibliographystyle{icml2020}
\clearpage
\appendix
\section{Policy Distillation Results}
\begin{figure}[h]
\centering
\begin{subfigure}{0.8\columnwidth}
\includegraphics[width=\textwidth]{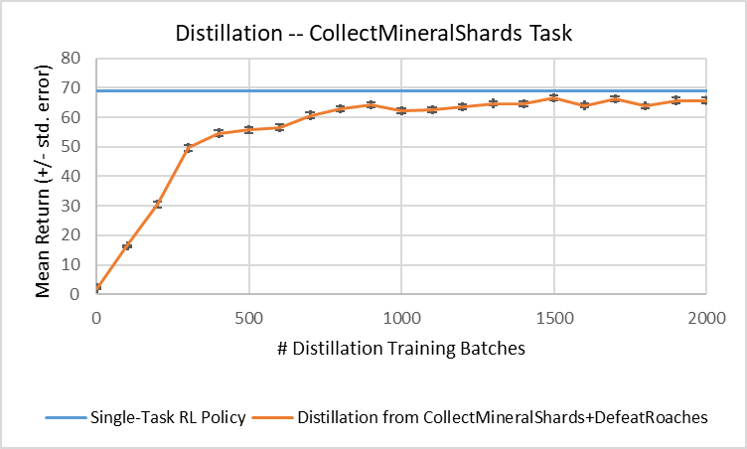}
\caption{}
\label{fig:sc2-forward-distillation:collectmineralshards}
\end{subfigure}

\begin{subfigure}{0.8\columnwidth}
\includegraphics[width=\textwidth]{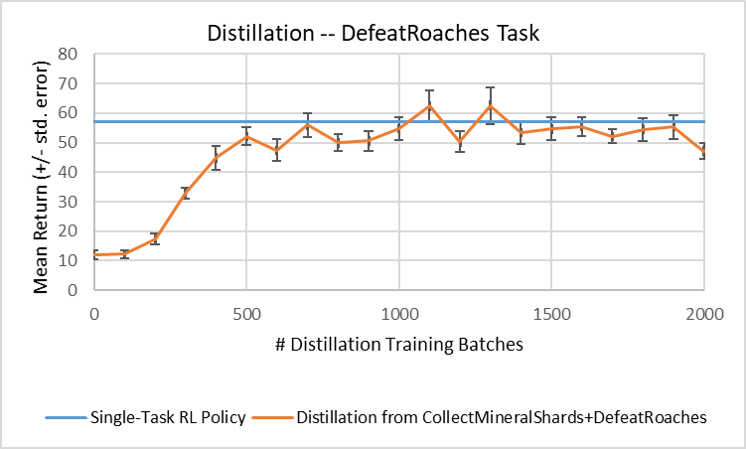}
\caption{}
\label{fig:sc2-forward-distillation:defeatroaches}
\end{subfigure}
\caption{Policy distillation}
\label{fig:sc2-distillation}
\end{figure}

We use a policy distillation approach for policy consolidation. As a proof-of-concept, we conducted an experiment using policy distillation to combine two SC2 policies -- for the CollectMineralShards and DefeatRoaches tasks -- into a single policy. In the experiment, we use real observations from both tasks for distillation, rather than sampling one set of observations from a generative model. Figure~\ref{fig:sc2-distillation} compares the performance of the distilled policy to the performance of single task policies. Distillation is able to preserve the control knowledge embodied in both policies while compressing them into a single policy, and about 100x fewer training batches are required for distillation than were required originally to learn the policies being distilled. Policy distillation thus provides an effective and efficient means of knowledge consolidation for our framework.

\section{Supplementary Experimental Results}
\begin{figure*}[h]
    \centering
    \hfill
    \begin{subfigure}{.32\textwidth}
    \centering
    \includegraphics[width=\linewidth]{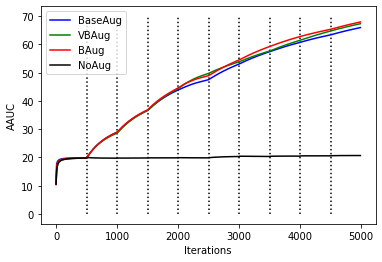}
    \caption{Average area under the curve averaged over classes vs iterations.}
    \label{fig:aauc}
    \end{subfigure}%
    \hfill
    \begin{subfigure}{.32\textwidth}
    \centering
    \includegraphics[width=\linewidth]{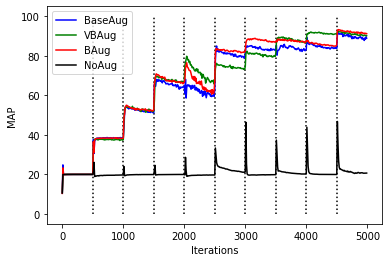}
    \caption{Mean average precision vs iterations.}
    \label{fig:map}
    \end{subfigure}
    \hfill
    \begin{subfigure}{.32\textwidth}
    \centering
    \includegraphics[width=\linewidth]{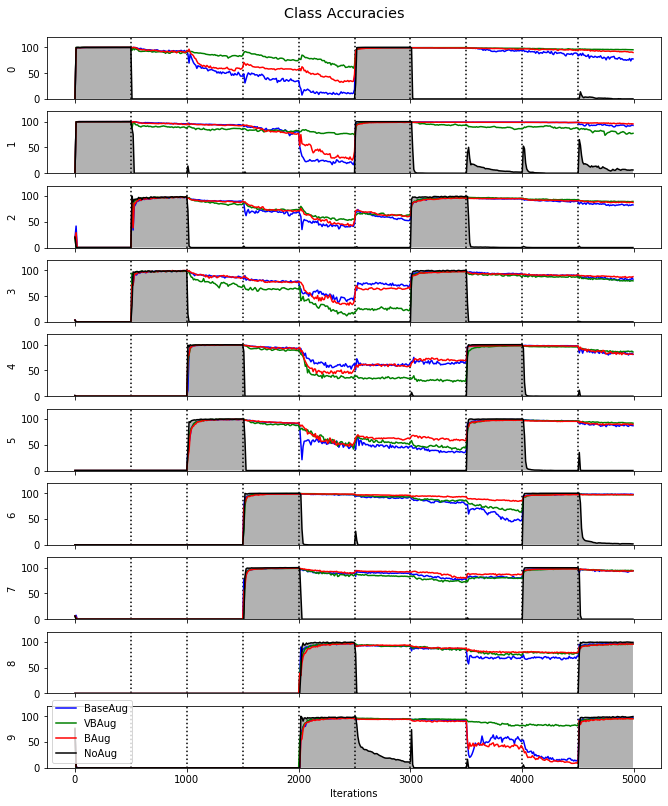}
    \caption{Per-class (y-axis) accuracy over iterations (x-axis).}
    \label{fig:class_acc}
    \end{subfigure}%
    \hfill
    \caption{SplitMNIST: Per-task breakdown of accuracy vs iterations. Each task is seen for 500 iterations. Steady increase in these metrics indicates successfull continual learning without forgetting.}
    \label{fig:fig}
\end{figure*}

\begin{figure*}[h]
    \centering
    \hfill
    \begin{subfigure}{.32\textwidth}
    \centering
    \includegraphics[width=\linewidth]{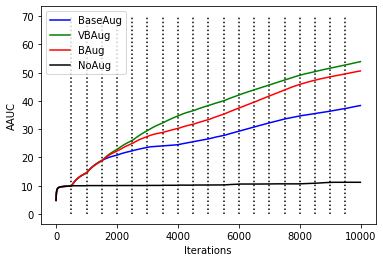}
    \caption{Average area under the curve averaged over classes vs iterations.}
    \label{fig:aauc_fm}
    \end{subfigure}%
    \hfill
    \begin{subfigure}{.32\textwidth}
    \centering
    \includegraphics[width=\linewidth]{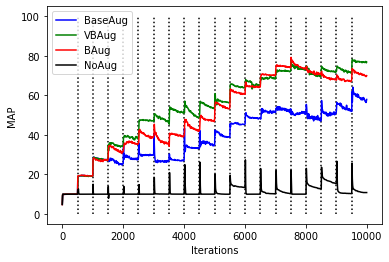}
    \caption{Mean average precision vs iterations.}
    \label{fig:map_fm}
    \end{subfigure}
    \hfill
    \begin{subfigure}{.32\textwidth}
    \centering
    \includegraphics[width=\linewidth]{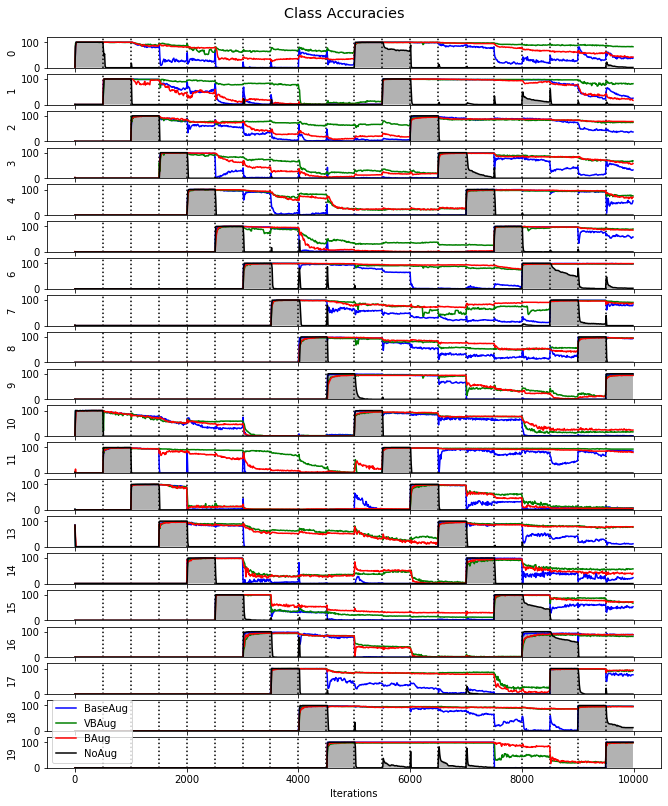}
    \caption{Per-class (y-axis) accuracy over iterations (x-axis).}
    \label{fig:class_acc}
    \end{subfigure}%
    \hfill
    \caption{Split(MNIST+FashionMNIST): Per-task breakdown of accuracy vs iterations. Each task is seen for 500 iterations. Steady increase in these metrics indicates successfull continual learning without forgetting.}
    \label{fig:fig_fm}
\end{figure*}

\begin{table*}[h] \small
    \begin{tabular}{cccccc}
    \hline
    BuildMarines & CollectMineralShards & DefeatZerglingsAndBanelings & MoveToBeacon & CollectMineralsAndGas & DefeatRoaches \\\hline
    \includegraphics[width=0.10\textwidth]{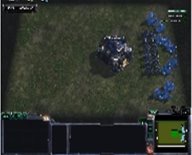} &
    \includegraphics[width=0.10\textwidth]{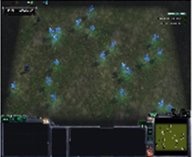} &
    \includegraphics[width=0.10\textwidth]{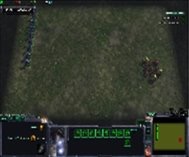} &
    \includegraphics[width=0.10\textwidth]{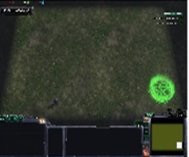} &
    \includegraphics[width=0.10\textwidth]{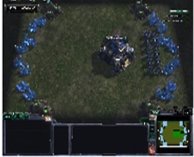} &
    \includegraphics[width=0.10\textwidth]{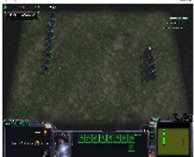} \\
    \includegraphics[width=0.10\textwidth]{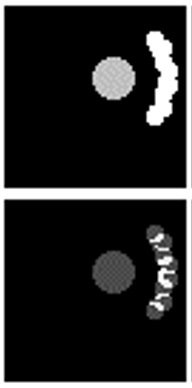} &
    \includegraphics[width=0.10\textwidth]{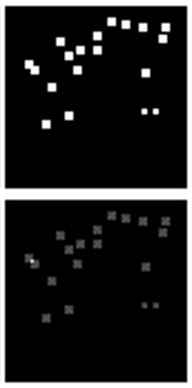} &
    \includegraphics[width=0.10\textwidth]{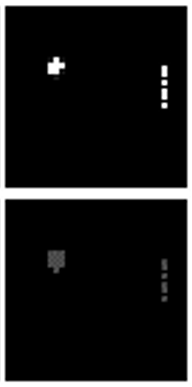} &
    \includegraphics[width=0.10\textwidth]{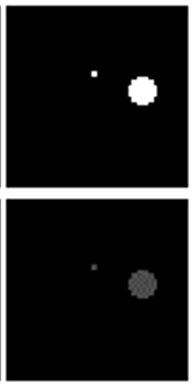} &
    \includegraphics[width=0.10\textwidth]{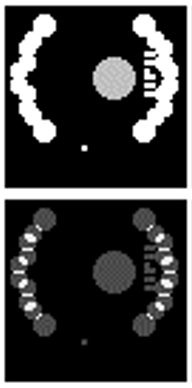} &
    \includegraphics[width=0.10\textwidth]{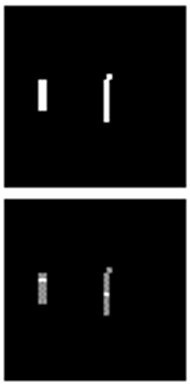} \\\hline
    \end{tabular}
    \caption{SC2 Data Representation}
    \label{tab:SC2_data}
\end{table*}

\begin{figure*}[h]
    \centering
    \includegraphics[width=0.9\linewidth]{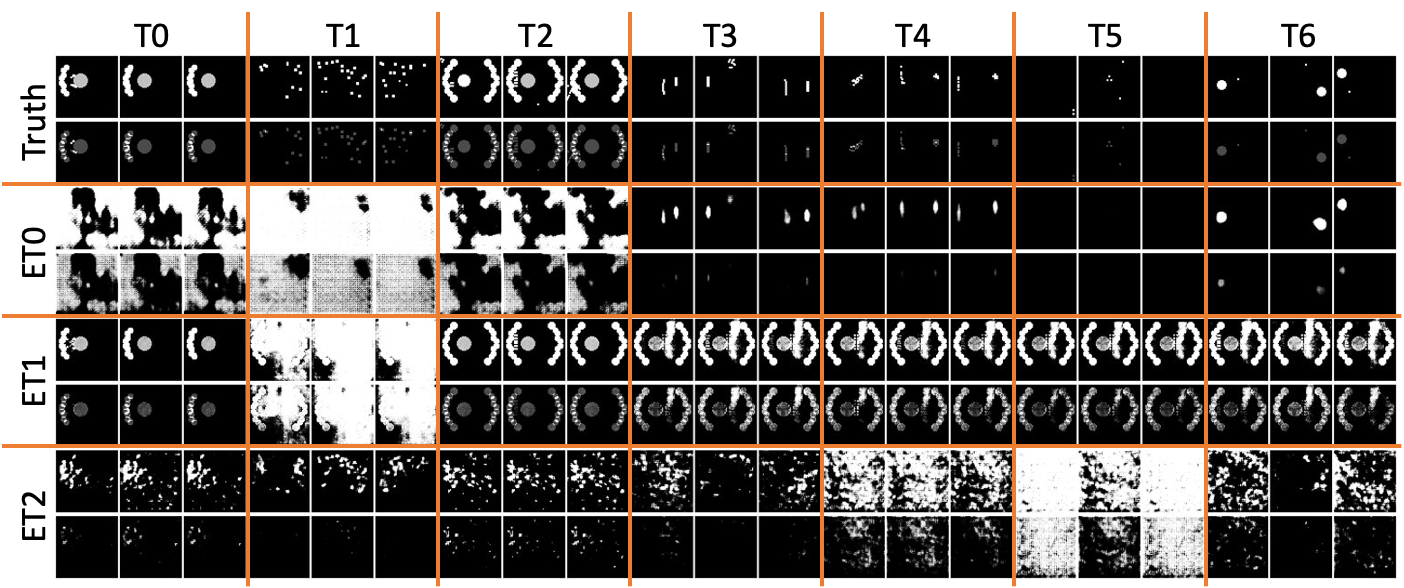}
    \caption{OWVAE reconstructions for SC2 tasks.}
    \label{fig:reconstruction_sc2}
\end{figure*}

\end{document}